\documentclass[fontsize=12pt]{scrartcl}

\usepackage[utf8]{inputenc} 
\usepackage{authblk}
\usepackage[colorlinks=true]{hyperref}
\hypersetup{
    colorlinks,
    citecolor=black,
    filecolor=black,
    linkcolor=black,
    urlcolor=black
}
\usepackage{tabularx}
\usepackage{booktabs}
\usepackage{caption}
\usepackage{graphicx}
\usepackage{float}
\usepackage[font=bf,labelfont=bf]{caption}
\usepackage{amsmath}
\usepackage[sfdefault]{roboto}
\usepackage{setspace}

\begin{document}

\title{\vspace{-2em} \textmd{Automatic detection of problem-gambling signs from online texts using large language models}}
\date{\small \today}

\author[1*]{Elke Smith}
\author[2]{Nils Reiter}
\author[1]{Jan Peters}
\renewcommand\Affilfont{\small}
\affil[1]{\small Department of Psychology, Biological Psychology, University of Cologne, Germany}
\affil[2]{\small Department of Digital Humanities, University of Cologne, Germany}

\maketitle
\thispagestyle{empty}

\onehalfspacing

\begin{abstract}
Problem gambling is a major public health concern and is associated with profound psychological distress and economic problems. There are numerous gambling communities on the internet where users exchange information about games, gambling tactics, as well as gambling-related problems. Individuals exhibiting higher levels of problem gambling engage more in such communities. Online gambling communities may provide insights into problem-gambling behaviour. Using data scraped from a major German gambling discussion board, we fine-tuned a large language model, specifically a Bidirectional Encoder Representations from Transformers (BERT) model, to predict signs of problem-gambling from forum posts. Training data were generated by manual annotation and by taking into account diagnostic criteria and gambling-related cognitive distortions. Using k-fold cross-validation, our models achieved a precision of 0.95 and F1 score of 0.71, demonstrating that satisfactory classification performance can be achieved by generating high-quality training material through manual annotation based on diagnostic criteria. The current study confirms that a BERT-based model can be reliably used on small data sets and to detect signatures of problem gambling in online communication data. Such computational approaches may have potential for the detection of changes in problem-gambling prevalence among online users.
\end{abstract}

\vspace{3em}
\footnoterule \vspace{0.5em} \small \textsuperscript{*} Corresponding author. Email: \href{mailto:e.smith@uni-koeln.de}{e.smith@uni-koeln.de}

\section{Introduction}

Gambling refers to the act of betting or wagering on an event with an uncertain outcome, with the primary intent of winning money or material goods. This activity is usually associated with games of chance, such as slot machines, lotteries, and card games. In contrast to gaming, which contains elements of strategy or competition, the gambling outcomes are primarily determined by chance. In some individuals, however, gambling behaviour can escalate and result in a loss of control over gambling behaviour. Disordered gambling can then lead to severe psycho-social consequences. Affected individuals typically feel an urge to gamble and are preoccupied with gambling, often have difficulties to control or stop gambling, gamble high amounts of money and try to compensate for losses by further gambling \cite{american2014diagnostisches}. Disordered gambling commonly leads to psychosocial distress, financial problems, problems at work and in the family, and heightened levels of stress \cite{fong2005biopsychosocial}. In addition to escalating gambling behaviour and resulting problems such as financial difficulties, disordered gambling is linked to specific gambling-related cognitive distortions, i.e. irrational and superstituous cognitions. For example, individuals overestimate their chances of winning, downplay the risks associated with gambling and hold maladaptive and erroneous beliefs about the role of skill in gambling (illusion of control) \cite{goodie2013measuring,johansson2009risk}.

Disordered gambling is a mental disorder listed in the Diagnostic and Statistical Manual of Mental Disorders (5th ed., DSM-5) \cite{american2014diagnostisches}, and referred to as pathological gambling in the International Classification of Diseases (11th revision, ICD-11) \cite{world2018icd}. Prevalence rates for disordered gambling vary from country to country, but figures between 0.1 and 0.6 percent are reported for Europe \cite{griffiths2010problem}. The condition is strongly comorbid with a range of psychiatric symptoms, including anxiety, depression and substance use \cite{sundqvist2019problem}, making it a major public health concern \cite{lancet2017problem}.

In Germany, land-based (terrestrial) gambling is subject to several regulations and restrictions such as mandatory programmes for self-exclusion, age and marketing restrictions \cite{gluestv,gluestv21}. Following a revision of the German State Treaty on Gambling in 2021 (GlüStV 2021), online gambling was legalised in 2021 \cite{gluestv21}. Gambling behaviour and gambling-associated problems are often identified through self-report, including interviews and questionnaires. These, in turn, are subject to known biases such as social desirability \cite{krumpal2013determinants} and recall bias \cite{bradburn1987answering}, which describe the tendency of respondents to give answers that are assumed to be viewed as favourable by others, and the inaccurate recall of past events, respectively. Thus, more direct measures of behaviour might be more informative in some cases. However, a direct assessment of gambling behaviour is technically challenging and would require online tracking or on-site observations and provider cooperation.\newline

\subsection*{Problem gambling and participation in online communities}
On the internet, there are numerous gambling communities and online discussion boards where users exchange information on gambling experiences, putative strategies and gambling-related problems \cite{griffiths2010use,sirola2021online}. Survey studies report that individuals who exhibit higher levels of problem gambling engage more in such communities \cite{sirola2018excessive,sirola2019loneliness}. For instance, in a Finnish survey, more than half of online gambling forum users reported some problems with gambling \cite{sirola2018excessive}, as measured by the South Oaks Gambling Screen (SOGS) \cite{lesieur1987south}. This highlights the importance of understanding the experiences and problems related to gambling that are shared in these online forums. The study of online gambling communities may therefore be a fruitful way to provide further insights into gambling behaviour and associated problems \cite{griffiths2010use}. Since research suggests a role of online communities in the development and persistence of problem gambling, monitoring online gambling communities could further be useful in initiating low-threshold preventive measures \cite{sirola2018excessive,sirola2019loneliness}. There are both qualitative and quantitative approaches, and combinations of both, to study content from online communities \cite{caputo2015sharing,griffiths2010use,im2006online}. Given the large amounts of text-based information contained in these communities, computational methods for information extraction have been widely adopted, for instance, to predict mental health conditions by using social media posts \cite{chancellor2020methods,merchant2019evaluating}.

\subsection*{Computational linguistics in clinical psychology}
To derive information with potential clinical or public health implications from large bodies of text, e.g. with respect to clinical psychological questions, computational linguistics approaches can be helpful. The application of natural language processing (NLP) methods in clinical settings may facilitate dealing with big data, for instance by retrieving infomation with potential clinical or public health implications, and by structuring and synthesising information from clinical documents and scientific literature \cite{garner2004engineering,neveol2018clinical}. For instance, performing computerised lexical analysis of narratives from individuals with gambling or buying compulsions with software for Linguistic Inquiry and Word Count (LIWC) \cite{pennebaker1999linguistic} revealed differences in emotional tone and authenticity between compulsive buying and gambling narratives, suggesting that differences in phenomenology may be automatically identified from written narratives.

\subsection*{Automatically predicting signs of problem gambling}
Automatically detecting problem-gambling content, indicating potentially problematic gambling behaviour and early signs of gambling addiction from online texts could aid website operators in their monitoring activities in the context of player protection. Such approaches may also provide researchers with tools for measuring changes in online posting behaviour that might be related to specific events such as lockdowns following the COVID-19 pandemic in 2020, or the introduction of the GlüStV 2021 in Germany \cite{gluestv21}. To date, few studies have specifically addressed the issue of automatically detecting signs of problem gambling from online posts.

There is some published work from the 2021 and 2022 editions of the Early Risk Detection on the Internet (eRisk) workshop \cite{parapar2021overview,parapar2022overview}. The eRisk workshop contributes to developing and evaluating computational methods to detect risk factors for mental health problems on the internet, including depression and problem gambling. Relying on texts from Reddit, and following a user-centred approach, users who had posted or commented in subreddits dealing with problem gambling or gambling addiction were considered as users at-risk, while control users were either taken from the publicly available Reddit Self-reported Depression Diagnosis (RSDD) \cite{yates2017depression} and eRisk 2018 depression \cite{losada2016test} datasets, or obtained by selecting users which had posted in gambling-unrelated subreddits. 

Evaluating machine learning classification performance is typically based on precision, recall, and F1 score. Precision denotes the fraction of target class elements among the retrieved elements, while recall (also referred to as sensitivity) denotes the fraction of target class elements that were retrieved. The F1 score is a combined measure of precision and recall, specifically the harmonic mean between the two. The metrics range between 0 and 1, with higher scores indicating better classification performance. Training different machine learning models at eRisk \cite{parapar2021overview,parapar2022overview} for predicting early signs of problem gambling, yielded F1 scores of 0.72 \cite{loyola2021unsl} and 0.87 \cite{fabregat2022uned}, when using a BoW representation with a support vector machine (SVM), and approximate nearest neighbour algorithm, respectively. Also, Bucur and colleagues fine-tuned a pre-trained Bidirectional Encoder Representations From Transformers (BERT) model to the task of predicting signs of problem gambling from gambling-related Reddit posts \cite{bucur2021early,bucur2022end}, yielding F1 scores of 0.27 and 0.41, respectively. While recall was high (0.98 and 0.99), precision was rather low (0.16 und 0.26). 

The aforementioned works demonstrate the feasibility of using machine learning algorithms and models, such as SVM and BERT, for automatically predicting signs of problem gambling. However, model performance was in some cases limited, thus leaving room for improvement, especially with regard to precision. Considering the potential applications of such models, e.g. for early detection of at-risk behaviours from social media, obtaining high precision is an essential requirement to avoid producing massive amounts of false positives. One potential reason for the low precision in previous work concerns the way in which the training material was constructed. The training data were not manually annotated, rather, all posts from the chosen subreddits were defined as positive class items, assuming that the majority of users posting there exhibit pathological gambling behaviour. In addition to concerns regarding validity, this may have led to significant numbers of posts from non-problem-gambling users being included in the positive (at-risk for problem gambling) class.

Building upon this, the aim of the present work was to automatically identify problem-gambling content in online forum posts using a pre-trained ML model for NLP. We expanded upon previous work by using manual annotation to define training labels. Annotation was based on both diagnostic criteria and gambling-related cognitions. Regarding diagnostic criteria, descriptions of one’s gambling behaviour in forum posts may provide information about whether an individual describes problems with gambling, such as gambling-related financial problems, urges to gamble or issues related to treatment and support. Regarding gambling-related cognitions, posts discussing gambling strategies or tactics might contain signatures of maldaptive beliefs (see above). To this end, posts scraped from a large German online gambling forum were manually annotated as containing problem-gambling content vs. containing gambling content only, based on standard diagnostic instruments for problem gambling. We then fine-tuning a pre-trained German version of the BERT model \cite{chan2019german} using the annotated forum posts to examine the degree to which such content types could be automatically detected.

\section{Materials and methods}

\subsection{Data collection and description}\label{sec:data}
The data used in the current work stem from a discussion board that is part of a German- speaking online casino and gambling website. The topics discussed centre around online casinos, slot machines, games such as roulette, blackjack and poker, and gambling addiction. The website was scraped using Python (version 3.6.9) \cite{python3}, and the Requests (version 2.18.4) and Beautiful Soup (version 4.6.0) \cite{richardson2007beautiful} libraries. In Germany, web scraping for scientific purposes is legal if no access restrictions are circumvented \cite{klawonn2019urheberrechtliche}. The data scraped from the website were openly accessible and technical measures designed to prevent web scraping were not disregarded. The scraped XML data were parsed into a relational SQLite database \cite{sqlite2020hipp} using the sqlite3 module for Python. The database contains the tables “post” and “author”. The 10 “author” table contains information about the author of a forum post, such as username, date of membership activation, and number of posts. In the context of this work, only data from the “posts” table was considered, which contains information about a forum post, including its publication date, URL (specifying the subforum), and its text. The database contains a total of 205,385 forum posts. The gambling addiction subforum contains 4,150 posts, 202 of which are initial posts. All other forum sections (excluding the gambling addiction subforum) contain 201,235 posts, including 7,705 initial posts (i.e., posts that are not replies to another post). The discussion board is grouped into eleven superordinate board topics (see Table \ref{tab:subforums}), with the highest number of posts found in the online casino subforum. Text length (tokens per post) is, on average, twice as high in the gambling addiction subforum (\textit{M} = 296.16) compared to all other subforums (\textit{M} = 141.54).

\begin{table*}[ht]
\setcapindent{0pt}
\caption{\raggedright Distribution of posts per subforum}
\label{tab:subforums}
\begin{tabular}{lll}\\\toprule  
Subforum & \textit{N} posts \\
\midrule
Rules and guidelines & 9349 \\
Blackjack & 308 \\
Poker & 296 \\
Roulette & 614 \\
Other games of chance & 4701 \\
Slot machines and slot games & 6079 \\
Gambling arcades and casinos & 3365 \\
Casino complaints & 14328 \\
Gambling addiction & 4150 \\
Online casinos & 140818 \\
Miscellaneous & 21377 \\
\bottomrule
\end{tabular}
\end{table*}

\subsection{Selection of gambling forum posts}
Since BERT has a maximum length limit of 512 tokens (see also section \ref{sec:model}), only posts with length $\leq$ 512 tokens were considered for annotation. Using all posts and automatically truncating them before classification would result in the annotated data not being comparable to the training data. It is conceivable that a post contains descriptions after the 512th token that make it a target. Only initial posts (i.e., posts that are not replies to other posts) were considered for annotation. After removing all posts with a token length $>$ 512, 168 initial posts remained from the addiction subforum, and 7466 initial posts from all other subforums. The gambling addiction subforum potentially contains more target (i.e. problem gambling) posts, whereas posts from all other subforums potentially contain more non-target (i.e. gambling only) posts. Since the gambling addiction subforum contains few initial posts, all posts from this forum were exported for manual annotation (\textit{N} = 168). Twice as many posts were exported from all other subforums (\textit{N} = 336, randomly selected). However, text length (in tokens, as determined through tokenisation with the NLTK Tokenizer package \cite{bird2009natural}) was, on average, higher in the gambling addiction subforum compared to all other subforums (see section \ref{sec:data}), even after removing posts with token length $>$ 512. To prevent the classifier from using text length as a criterion, exporting posts from all non-gambling addiction subforums was performed with weighted random sampling based on text length to obtain similar distributions in terms of token length. The difference in token length between posts exported from the addiction subforum and the random weighted sample of posts from all other subforums was not significantly different (\textit{T} = -0.99, \textit{p} = 0.32, Welch’s t-test). To prevent the annotation from being influenced by prior knowledge about the subforum a post came from, a single post was exported with its text content and unique ID only, ensuring blindness to the post origin.

\subsection{Annotation of gambling forum posts}
To create training material for the classifier, an annotation guide was developed for labelling forum posts as containing problem-gambling content (target posts), or containing gambling content only (non-target posts). Posts that could not be classified were labelled “inconclusive”. The guide consists of annotation instructions, as well as an annotation form in which the applicable criteria for a single forum post may be noted (see Supporting Information S1). The annotation guide is based on the criteria for gambling disorder as listed in the DSM-5 \cite{american2014diagnostisches} and on the items from the Gambling Related Cognitions Scale (GRCS) \cite{raylu2004gambling}. The focus was put on coding the presence/absence of gambling-related problems and/or endorsement of gambling-related cognitions (i.e. target vs. non-target classification), rather than on a continuous measure of e.g. gambling severity. The temporal dimension of pathological gambling according to DSM-5 criteria (i.e., meeting a specific number of criteria within twelve months) was not taken into account in the annotation, as this cannot be reliably assessed on the basis of individual posts. Problem gambling is often accompanied by gambling-related cognitive distortions, i.e. erroneous beliefs about gambling, such as the ability to predict or control gambling outcomes \cite{johansson2009risk,raylu2004gambling}. To this end, the GRCS, a reliable and valid scale for the assessment of such beliefs, was additionally considered during annotation. The GRCS is a questionnaire developed to identify gambling-related cognitions in individuals that engage in gambling \cite{raylu2004gambling}. The instrument contains 23 items representing the 5 subscales gambling expectations, illusion of control, predictive control, inability to stop gambling, and interpretive bias. For annotation purposes, the criteria of the DSM-5 and items of the GRCS were categorised into the three subdomains (1) pathological gambling, (2) gambling-related problems, and (3) gambling-related cognitive distortions. Items from both instruments covering similar characteristics were grouped together. Since the focus was on detecting problem-gambling content in individual posts, a post was considered a target post (problem-gambling-related) if it contained at least one statement about the presence of pathological gambling behaviour, gambling-related problems, or gambling-related cognitive distortions. It was deemed irrelevant, whether the descriptions related to the author of a post or a related person (i.e. a friend or relative), or whether the author described current or past gambling-related problems. A post containing no such descriptions, but descriptions of gambling-related topics (e.g., casino complaints), was considered as non-target. Posts were also coded as targets if none of the items were explicitly reported in the descriptions, but the person clearly self-identified as being addicted to gambling, was actively seeking help or treatment for gambling disorder, or mentioned currently undergoing treatment.

\begin{figure}[H]
\captionsetup{format=plain,labelfont=bf}
\caption{\raggedright {Categories assigned during annotation of the forum posts}}
\fontsize{9pt}{9pt}\selectfont
In total, 504 posts were coded, yielding 348 posts labelled as gambling, 11 posts labelled as inconclusive (not included in the training set), and 138 posts labelled as problem gambling (PG). Seven posts were excluded for being unrelated to gambling (not depicted here). The problem gambling set contained 114 posts with signs of problem PG behaviour, 70 posts with signs of gambling-related (GR) problems, and 23 posts with GR cognitive distortions. Note that the three problem gambling subdomains are not mutually exclusive and therefore do not sum to the total number of PG posts.
\label{fig:annotation} \\
\begin{center}
\includegraphics[width=0.6\textwidth]{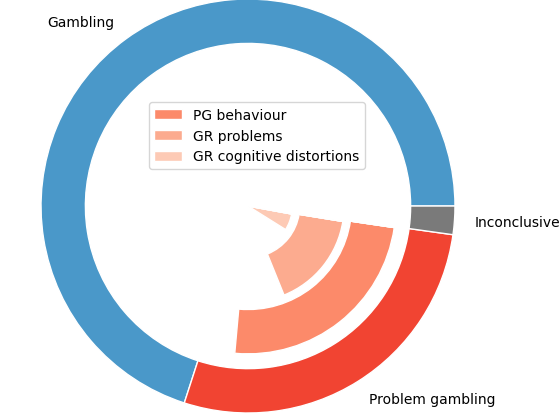}
\end{center}
\end{figure}

In total, 504 posts were coded by manual and blind annotation using the annotation guide (see Figure \ref{fig:annotation}). Seven posts were considered neither as target nor non-target posts and not included in the training set (one empty post, two posts from which the text content had been deleted by an administrator, and four posts that were not from regular users, but advertisements from universities to recruit study participants in the context of gambling addiction research). 11 posts were labelled as inconclusive and also not included in the training set, since it was not clear from the descriptions whether problem-gambling behaviour or cognitions were described. This was the case for posts in which persons asked questions about gambling addiction in a way that suggested they may be related to themselves, but without being specific about themselves, or posts in which the connection between described problems and gambling remained vague. 138 posts were annotated as targets, i.e., containing problem-gambling content, and 348 posts non-targets, i.e., containing other gambling content. Among problem-gambling posts, frequently mentioned issues were financial difficulties as well as repeated unsuccessful attempts to stop gambling. Gambling-related cognitive distortions often manifested as the belief that specific numbers or changing numbers relates to the chances of winning (illusion of control), and the belief in hot streaks (predictive control). Gambling posts often contained questions about the trustworthiness of specific casinos or complaints about casinos (mostly problems with payouts after winnings). Examples of annotated target, non-target and inconclusive posts, respectively, are provided as Supporting Information (S2).

\subsection{Model description}\label{sec:model}

Bidirectional Encoder Representations from Transformers (BERT) is a pre-trained transformer-based ML model developed by the Google AI Language team in 2018 \cite{devlin2018bert}. Pre-trained language models are trained on a huge corpus, thereby learning universal language representations, and then fine-tuned to a specific task in a subsequent step. For the present work, BERT was fine-tuned to the task of classifying forum posts as containing problem-gambling content vs. gambling content only. A large body of work has demonstrated that pre-trained language models outperform state-of-the-art models, such as recurrent neural networks, in a wide range of NLP tasks, and that these significantly reduce the amount of required training data (see, e.g., \cite{dai2019transformer,devlin2018bert,lakew2018comparison,wolf2020transformers}). BERT enables solving NLP tasks in a supervised fashion when the dataset labelled for training is not large enough for achieving satisfactory classification performance using a model trained from scratch. Being a transformer model, BERT is based on the concept of self-attention. Attention describes a mechanism that calculates the importance of tokens relative to other tokens in a sequence (e.g., words in a sentence), or the likelihood that tokens appear together. For a detailed description of the model architecture, the reader is referred to \cite{vaswani2017attention}.

\subsection{Model implementation}

BERT was implemented with Python (version 3.10.6) and the machine learning framework PyTorch (version 1.12.0) \cite{NEURIPS2019_9015}, and the Transformers library (version 4.21.0). For the current task, the German language model bert-base-german-uncased \cite{bertbasegermanuncased} was used. The German BERT model is based on BERT\textsubscript{BASE} (12 encoder layers, 768 hidden units, 12-self-attention heads, 110 m parameters) and has been pre-trained on a German Wikipedia and OpenLegalData dump and news articles. For fine-tuning, a regression head (linear layer) was added on top of the output by implementing BertForSequenceClassification \cite{bertforsequenceclassification}. The maximum input sequence length was set to 512 tokens (maximum possible sequence length of the BERT model). Shorter texts were padded up to the maximum sequence length.

\subsection{Preprocessing and training data size}
All input data were lowercased when tokenising using the BertTokenizer. Classification performance was then compared for a dataset that was not preprocessed further and a dataset from which punctuation characters were removed. This was done since punctuation in conjunction with semantic content may provide additional information related to e.g. emotional tone and urgency of the described issues. To consider the influence of the training data size, the model was fine-tuned using different set sizes. Specifically, the model was fine-tuned using one-half, two-thirds, and all target items, plus the same number of non-target items, respectively, yielding set sizes of 69/69, 92/92, and 138/138 (\textit{N} non-target or gambling items, \textit{N} target or problem-gambling items). When using subsamples of the annotated dataset (which contained more non-target than target items) the items were sampled with weighted subsampling based on text length to ensure that text length in tokens was not significantly different between target and non-target items (assessed with Welch’s t-tests for each dataset, with \textit{p} $<$ .05 for all tests). The model was also fine-tuned using all annotated items, which resulted in an imbalanced data set of size 348/138. Further, the model was fine-tuned using upsampling to match the number of problem-gambling items to the number of gambling items, yielding a balanced data set of size 348/348. For this purpose, the texts were converted to a matrix of term frequency-inverse document frequency (TF-IDF) features (a measure for the relevance of a word to a text document) in and then upsampled using the Synthetic Minority Over-Sampling Technique (SMOTE), as implemented in Imbalanced-Learn \cite{imblearn}.

\subsection{Fine-tuning and performance assessment}
The choice of training hyperparameters was guided by the values reported as optimal for fine-tuning the original BERT model \cite{devlin2018bert} and on the values reported in Bucur and colleagues for fine-tuning BERT to problem-gambling subreddits \cite{bucur2021early,bucur2022end}, yielding a batch size of 16, using the Adam optimiser with weight decay (epsilon = 1e-08) for 2 epochs with a learning rate of 5e-5. Training was performed under Ubuntu (version 18.04.6 LTS) on a Dell Precision Workstation 5820 using a Nvidia Quadro RTX 5000 GPU. Model performance was evaluated based on average accuracy, precision, recall, and F1 scores. While accuracy measures the fraction of correct predictions, it is not suitable for imbalanced data sets and does not allow for a comprehensive evaluation of the type of errors made. Precision may be considered a measure of quality, models with high precision may miss positive instances. Recall (also referred to as sensitivity) may be considered a measure of quantity, models with high recall tend to classify instances as belonging to the target class, thereby producing more false positives. Since, increasing precision usually comes at the cost of reduced recall, and vice versa, the F1 score takes into account both precision and recall. Performance was assessed for each fold using k-fold cross-validation with \textit{k} = 5. This way, the data set was randomly split into 5 sub sets. In five turns, four of those were used for training, while the fifth was held out for testing.

\section{Results}

\subsection{Model validation}
\begin{figure}
\captionsetup{format=plain,labelfont=bf}
\caption{Accuracy (A), precision (B), recall (C) and F1 scores (D) for each fold (points) and average accuracy with standard deviations across folds (bars), for each training set size and preprocessing pipeline, sorted by the number of problem-gambling posts.}
\fontsize{9pt}{9pt}\selectfont
The model was fine-tuned using one-half, two-thirds, and all target items, plus the same number of non-target items. The first number in the training set size denotes the count of non-target/gambling posts, the second number denotes the count of target/problem-gambling posts. The data with a training set size of 348/348 contain upsampled problem-gambling posts. For reasons of completeness, accuracy is depicted for all model runs, note, however, that accuracy is an inappropriate measure for unbalanced datasets (348/138). Following \cite{roder2018gerbil}, precision, recall and F1 are zero, when a processed batch contains no target (i.e. problem-gambling post), but the classifier returns one or more targets.
\label{fig:metrics} \\
\begin{center}
\includegraphics[width=\textwidth]{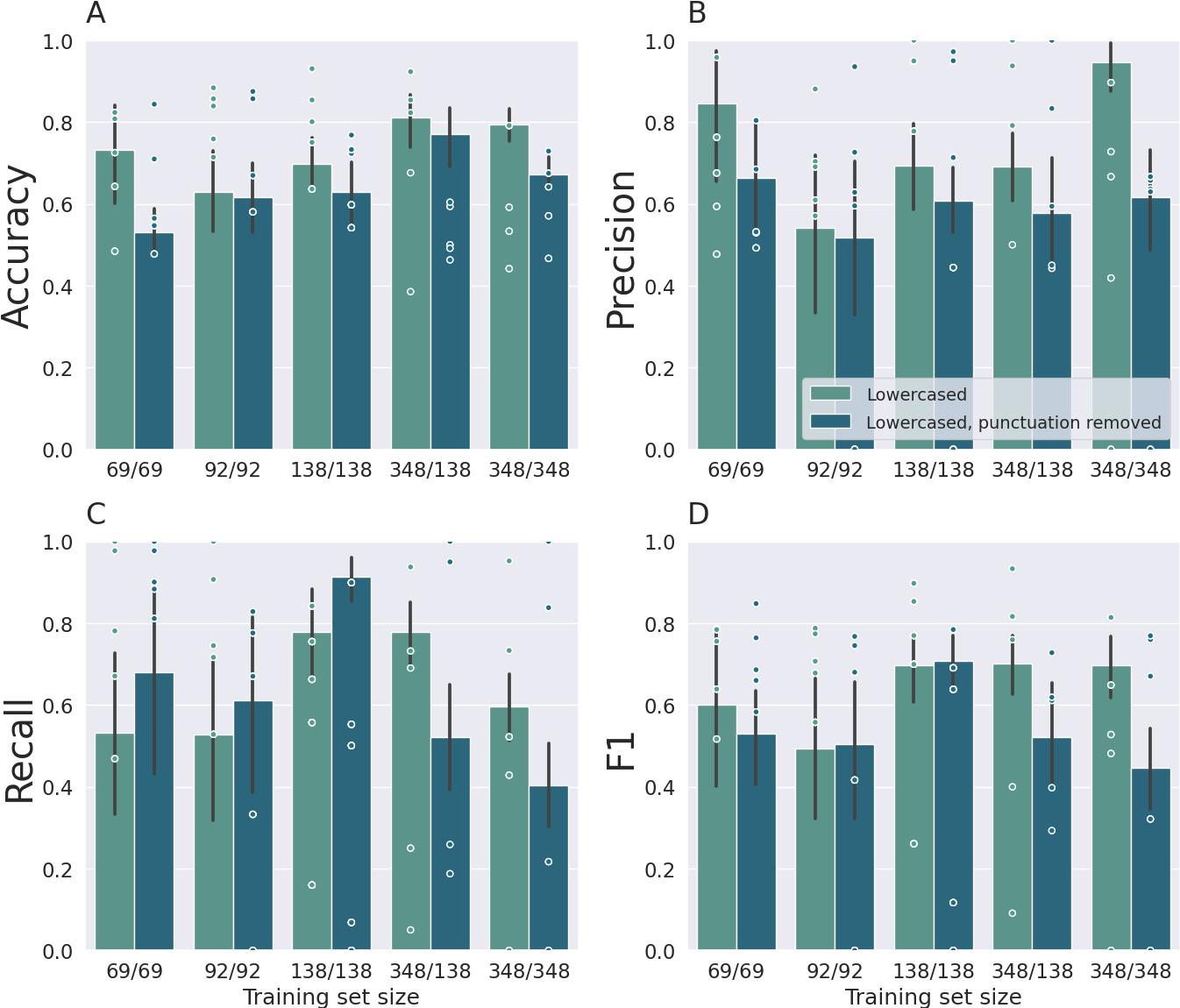}
\end{center}
\end{figure}

Prediction performance of the model was compared based on the average performance scores across folds for each preprocessing pipeline and training set size (see Table \ref{tab:metrics} and Figure \ref{fig:metrics}). Performance was above the baseline for binary classification (above 50\%, for the balanced data sets only) for all models with lowercasing only (keeping punctuation) and a training set size of 138/138 and larger. Accuracy, precision and F1 scores were higher for the models trained on larger training set sizes (see Figure \ref{fig:metrics}, panel A, B and D). The best model according to the F1 score proved to be the model using data with lowercasing and punctuation removal and a training set size of 138/138 (see Table \ref{tab:metrics}, and Figure \ref{fig:metrics}, panel D). Accuracy and precision increased with training set size (see Figure \ref{fig:metrics}, panels A and B), while recall did not systematically vary with training set size. Precision and recall varied substantially between folds (see standard deviations in Figure \ref{fig:metrics}, panels B and C). The highest precision was found for the lowercased training data set with a training set size of 348/348 (see Figure \ref{fig:metrics}, panel B). Overall, keeping punctuation improved predictive performance. Upsampling problem-gambling posts (set size of 348/348) yielded an improvement in accuracy and precision, but not recall, compared to the balanced dataset (set size of 138/138) (see Figure \ref{fig:metrics}, panel A, B, and D).

\begin{table}[h]
\sffamily
\fontsize{11pt}{11pt}\selectfont
\caption{Classification performance (means and standard deviations) for each preprocessing pipeline and training set size}
\vspace{0.5em}
\label{tab:metrics}
\begin{tabularx}{\textwidth}{lXXXXX}
\toprule
Preprocessing           & G/PG items     & Accuracy          & Precision       & Recall          & F1              \\
\midrule
Lowercasing             & 69/69          & 0.73 				(0.19) & 0.85 				(0.29) & 0.53 				(0.33) & 0.60 				(0.31) \\
                        & 92/92          & 0.63 				(0.20) & 0.54 				(0.39) & 0.52 				(0.38) & 0.49 				(0.34) \\
                        & 138/138        & 0.70 				(0.15) & 0.69 				(0.25) & 0.78 				(0.25) & 0.70 				(0.19) \\
                        & 348/138        & 0.81 				(0.18) & 0.69 				(0.26) & 0.78 				(0.24) & 0.70 				(0.22) \\
                        & 348/348        & 0.79 				(0.13) & 0.95 				(0.21) & 0.59 				(0.21) & 0.70 				(0.26) \\
Lowercasing 	and     	    & 69/69          & 0.53 				(0.09) & 0.66 				(0.2)  & 0.68 				(0.38) & 0.53 				(0.18) \\
punctuation 	removal	   & 92/92          & 0.62 				(0.17) & 0.52 				(0.37) & 0.61 				(0.43) & 0.50 				(0.33) \\
                        & 138/138        & 0.63 				(0.18) & 0.61 				(0.18) & 0.91 				(0.12) & \textbf{0.71}				(0.15) \\
                        & 348/138        & 0.77 				(0.23) & 0.58 				(0.40) & 0.52 				(0.39) & 0.52 				(0.31) \\
                        & 348/348        & 0.67 				(0.13) & 0.62 				(0.43) & 0.40 				(0.35) & 0.45 				(0.34)
\\ \bottomrule
\end{tabularx}
\vspace{0.1em} \\
\fontsize{9pt}{9pt}\selectfont
\parbox{\textwidth}{\textit{Note. }G: gambling, PG: problem gambling. Detection rates are reported as means across folds. The data set with a size of 348/348 contains upsampled target/problem-gambling posts. For reasons of completeness, accuracy is listed for all model runs, note, however, that accuracy is an inappropriate measure for unbalanced datasets (348/138). The best model in terms of F1 score is shown in bold.}
\end{table}

\subsection{Error analysis}
An error analysis was conducted for the best model according to the F1 score (training set size of 138/138, lowercasing and punctuation removal, see Table \ref{tab:metrics}). For this purpose, posts from all folds, for which the model’s predictions were false, were manually screened to identify potential systematic errors. For this model, recall was higher than precision (see Table \ref{tab:metrics} and Figure \ref{fig:metrics}, panel B and C). It was particularly noticeably that many of the false-positive predictions were made for posts including finance-related terms such as “money”, and “loss” (54\% of false-positive predictions), and “bank”, “bank account”, “bank details”, “account”, “account statement”, and “account number” (29\% of false-positive predictions). These posts were generally casino complaints (e.g. problems with payouts after winnings). This type of misclassification likely results from an overlap of content when writing casino complaints due to problems with payments, and content related to gambling-related financial problems, respectively. Further, many false-positive predictions were made for posts which included words such as “problem(s)” “help”, “support”, “warn” (91\% of false-positive predictions). These words also frequently appeared in problem-gambling posts, since users often posted to seek for help and to warn other users from slipping into disordered gambling. False-negative predictions (i.e., problem-gambling posts misclassified as gambling posts) were in some cases even made for posts containing terms that are clearly associated with problem gambling, such as “gambling addiction” (36\% of false-negative predictions) and “addiction” (43\% of false-negative predictions). Thus, a post was not classified as problem gambling based solely on the presence of a term such as gambling addiction. This may arise from the fact that only a small proportion of the posts annotated as problem-gambling posts actually contained such terms. In most cases, problem-gambling behaviour or gambling-related cognitive distortions were described without explicitly naming the condition.

Some of the screened posts for which erroneous predictions were made contained terms related to actual or predicted gambling-related cognitive distortions. Automatically detecting problem gambling by means of gambling-related cognitive distortions is not trivial, since it is partly a matter of rather subtle distorted cognitive processes that have to become apparent from the posts. Automatic detection of gambling-related cognitive distortions represents a challenge for the model and it further appears that the occurrence of financial terms in the context of casino complaints may have impaired the model’s performance since financial problems are a central characteristic of problem gambling and a frequently discussed topic in the annotated posts.

\section{Discussion}

The aim of the current work was to identify potential signatures of problem gambling behaviour, gambling-related problems and cognitive distortions in online forum posts. Using data scraped from a major German online gambling discussion board, a subsample of posts was manually annotated in consideration of the criteria for gambling disorder as listed in the DSM-5 \cite{american2014diagnostisches} and items of the GRCS \cite{raylu2004gambling}. The annotated data were then used to fine-tune a BERT model, a transformer-based pre-trained ML model for NLP \cite{devlin2018bert}, to the task of classifying posts into problem gambling (target category) vs. non-problem gambling content (non-target category).

\subsection*{Annotation of gambling forum posts}
The annotation revealed that problem-gambling posts constituted the minority class. This reflects the observation that the number of posts in the gambling addiction subforum made up a small proportion of the overall forum. Problem-gambling posts mostly contained explicit descriptions of problems related to gambling and symptoms of disordered gambling. Frequently mentioned topics were financial difficulties and repeated unsuccessful attempts to stop gambling. However, problem-gambling posts were not exclusively found in the problem-gambling subforum. In posts from the other subforums, excluding the casino complaints and rules and guidelines subforum, frequently discussed topics were gambling experiences and strategies. Since gambling is, by definition, based on chance, writing about gambling strategies may reveal gambling-related cognitive distortions of the user. This was the case, for example, when users assumed that specific strategies, colours or numbers influenced the course of the game, or when users assumed that the game was based on skill, rather than chance, or followed a predictable pattern. Some characteristics of gambling disorder following the DSM-5 criteria such as the duration of symptoms \cite{american2014diagnostisches} could of course not be reliably be determined from individual posts, and were therefore not considered. For future work, it might also be useful to code the severity of problem gambling content based on the number of described symptoms, and to annotate the posts using more granular categories, such as such as low-risk vs. high-risk problem gambling content for cases in which individuals describe following vs. not following self-imposed control strategies to regulate problem-gambling behaviour. Even if specific problem-gambling behaviour or symptoms are clearly reported, it must be kept in mind that this would obviously not suggest a clinical diagnosis. This requires a comprehensive structured clinical interview by a clinician. Furthermore, the level of analysis in the present approach is the individual post, and not the individual user. Needless to say, examining online posts cannot replace a comprehensive mental health assessment and diagnosis by an experienced practitioner \cite{karches2018against,hallowell2023democratising}.

\subsection*{Model performance}

Looking at the effects of training set size showed that performance was satisfactory for lowercased training data with 138 or more items per class. When training the model on additional, upsampled data, the model achieved high precision with a value of 0.95. Removing punctuations was not beneficial for model performance. On the contrary, it appears that punctuations conveyed some information relevant for distinguishing problem from non-problem gambling posts. This corresponds to the observation from the manual screenings that users often support descriptions of problems and negative emotions via punctuation \cite{hakami2022emoji,shoeb2019emotag}.

Bucur and colleagues \cite{bucur2021early,bucur2022end} took a similar approach to the one reported here, fine-tuning a pre-trained BERT model to detect signs of problem gambling. The authors scraped data from gambling addiction and problem-gambling subreddits, combining it with control data from other publicly available datasets. Comparing classification performance with regard to F1 score, the best model presented in the current work achieved a substantially better classification performance (yielding an F1 score of 0.71, compared to 0.27 and 0.41 \cite{bucur2021early,bucur2022end}). Two key differences may explain the differences in model performances. First, the training data from Bucur and colleagues \cite{bucur2021early,bucur2022end} were only loosely annotated (distant supervision), i.e. all posts from the chosen gambling addiction and problem-gambling subreddits were defined as positive class items, assuming that the majority of users posting in these subreddits exhibit problem gambling. The lack of manual annotation based on diagnostic criteria and gambling-related cognitive distortions in the construction of the training data can be seen critically in the context of construct validity and may have impaired prediction performance. Second, another key difference to the present work is the user-centred approach in collecting training data. From all unique users obtained from crawling the gambling-related subreddits, posts written in other, gambling-unrelated, subreddits were crawled to obtain both gambling-related and gambling-unrelated posts for the positive class. Negative class items were taken from control users from other openly available Reddit-based-datasets \cite{bucur2021early,bucur2022end}. By also including gambling-unrelated posts from the positive class users, the trained model may have missed important context or relevant information that could help distinguish signs of problematic gambling. Overall, the results of the current work demonstrate that by annotating forum posts based on diagnostic criteria and gambling-related cognitive distortions, satisfactory classification performance could be obtained with less training data compared to state-of-the-art work, using only 348 items per class as compared to processing a multiple of that (1,828 to 170,698 user writings, see \cite{bucur2021early,bucur2022end,fabregat2022uned,loyola2021unsl}).

While there is room for improvement in terms of recall, considering a scenario where signs of problem gambling shall be detected from a plethora of discussion forum or social media posts requires high precision to avoid producing large amounts of false positives. For instance, if website operators are to initiate certain action cascades upon detecting possible signs of problem gambling using a ML application (e.g., contacting a user), this would only be feasible for precise predictions. Having to process large amounts of false positive cases would require many resources, and moreover, could lead to users feeling disturbed or monitored when being contacted frequently. Critically, while the approach may be suitable for detecting early signs of problem gambling, it would be highly problematic if someone were to draw far-reaching conclusions from individual posts. Therefore, ethical and privacy aspects must always be taken into consideration when implementing such models.

\subsection*{Model error patterns}

To determine whether the errors that the model made in its predictions followed systematic patterns, an error analysis was conducted. Posts of the best model according to F1- score, for which false predictions were made during validation, were screened with regard to the (co)occurrence of specific terms. The error analysis essentially revealed that a substantial part of the posts for which false-positive predictions were made contained finance-related terms, and the word “problem(s)”. These types of gambling posts were mostly casino complaints featuring descriptions of problems with payouts at specific casinos. Likewise, problem-gambling posts frequently contained descriptions of (gambling-related) financial problems, which is why terms such as “problems” and “difficulties” were often used in combination with terms such as “bank” and “account balance”. The casino complaints subforum is the second largest subforum with regard to the absolute number of posts, and the overlap of terminology in problem-gambling posts and casino complaints posed a special challenge for the classifier. Considering the error patterns with regard to the transformer model architecture leads to revisiting the attention mechanism. The transformer model considers the relationships between tokens in an input sequence, including the context of the word “problem” when combined with finance-related words. The attention mechanism of the transformer model encodes a word based on other words in a sentence and indicates how strongly a word is associated with other words in the same sentence. Additionally, the positional encoding layer in both the encoder and decoders of the BERT model should account for the order of the tokens in the input sequence. This means that the model should be able to distinguish between different sequences of words, even if they contain the same words, based on their order in the input. However, it is possible that the model may still make errors in classifying text that contains the word “problem” in combination with finance-related words. One reason for this could be that the training data did not include enough examples of this specific pattern in casino complaints and problem-gambling posts, which may have led to the model not being able to learn the correct association between these words. Based on the described error patterns and the attention mechanism of the transformer model, it is possible that the model is over-emphasising the word “problem” and other finance-related terms in the input sequence while under-emphasising context (other tokens) that could help to differentiate between casino complaints and descriptions of gambling-related financial problems. To address this, it may be beneficial to add further training examples that reflect the nuances of the described problems more clearly.

Looking at false-negative predictions, it appeared that several posts were misclassified as gambling posts, even if they contained terms clearly associated with problem gambling, such as “gambling addiction” and ”treatment”. These were mainly posts in which the authors clearly described that they were addicted to gambling, while not mentioning many symptoms or gambling-related cognitions. That the presence of problem gambling “keywords” is not sufficient for the model to predict problem gambling, may be related to the observation that in the majority of problem-gambling posts, problem-gambling behaviours or related cognitive distortions were described without explicitly mentioning the condition.

\subsection*{Limitations and ethical aspects}
Overall, we used comparatively little data for classification, implying that a large part of the linguistic variance has not been explored. The implemented model is language-specific, and further, may not generalise to other platforms. Cross-validation was used to estimate how well the model would generalise. While overfitting may be detected using k-fold cross-validation, the performance estimates may still be overly optimistic if the model is overfitting. Different social media platforms may be associated with different posting behaviours and norms, which could affect the quality and representativeness of the data collected. The generalisability and applicability to texts from other online communities would have to be assessed in a next step. Further, the method for data collection may have introduced a sampling bias. Although survey studies indicate that individuals who demonstrate higher degrees of problem gambling tend to participate more actively in online gambling communities \cite{sirola2018excessive,sirola2019loneliness}, not all individuals who gamble participate in online gambling communities, and therefore, the training data may not be representative of the entire population of individuals who gamble.

A small fraction of posts was labelled as inconclusive and not included in the training set. In these cases, the descriptions provided were either to vague to classify the behaviour as either gambling or problem gambling, or users asked questions about problem gambling without being specific that the question related to themselves. The approach of not including inconclusive posts in the training data was taken to focus on clear and high-quality examples and to reduce noise since classification performance is strongly dependent on the quality and discriminatory power of the annotations and the amount of attribute noise \cite{zhu2004class}. However, excluding inconclusive posts may also have potential drawbacks. First, excluding posts limits the amount of available training data. Since only a small fraction of posts was labelled as inconclusive, this did not significantly impact on training set size. Second, the classifier may not perform as well in real-world scenarios containing inconclusive descriptions, which may mitigate its usefulness in practical applications. A potential outlook for further work could therefore be to include inconclusive posts as a third category for the machine learning classification. By creating a separate category for inconclusive posts, the classifier may be trained to distinguish between different types of posts, even if it cannot confidently assign them to the gambling or problem-gambling category. This may help to improve the robustness of the model for real-world scenarios. However, including a third category would also add complexity to the model and would require much more training data. Since inconclusive posts were comparatively rare, a substantially larger part of the data would have to be screened manually.

While modelling mental health status from social media usage may yield great benefits for monitoring site operators, some caution is also warranted when employing such techniques. Reviewing the literature, construct validity may frequently be questioned. It is not always clear, what the ground truth represents when annotation data with respect to signs of psychopathology \cite{merchant2019evaluating}. As an example, Vaishnavi and colleagues \cite{vaishnavi2022predicting} compared various ML-methods for the prediction of mental health, not indicating on what basis mental health was evaluated. In contrast, the annotation in the present study was performed manually based on clinically defined criteria, and we emphasise taking into account validated clinical criteria when implementing ML methods. Finally, considering such models for application in practice, it must be kept in mind that such methods bear the risk of algorithmic bias, i.e., systematic errors in machine-based decisions that create unfair outcomes for specific persons or groups \cite{delgado2022bias,walsh2020stigma}.

Using data obtained from public online forums offers the advantage of assembling large datasets without any interference from researchers, avoiding effects of direct observation. However, this method also requires significant ethical considerations. The role of "perceived privacy" within online communities should be considered \cite{eysenbach2001ethical}. For instance, if access to posts is restricted to registered users, it implies that contributors may perceive their communication as occurring in a private realm. If platform providers implement technical measures to limit data access, researchers should not circumvent these measures \cite{landers2016primer}. Adhering to these principles, we gathered publicly available data without bypassing any technical barriers.

\subsection*{Perspectives}

Problem gambling is a public health concern and leads to serious psychological and economic consequences, such as depression and indebtedness for many of those affected \cite{fong2005biopsychosocial}. According to a survey by the Federal Centre for Health Education in Germany (Bundeszentrale für gesundheitliche Aufklärung, BZgA) more than 400,000 individuals are estimated to exhibit problematic or pathological gambling behaviour in Germany in 2019 \cite{banz2019glucksspielverhalten}. The increased availability of online gambling services and the legalisation of online gambling in more and more countries, including Germany, may pose a high risk for vulnerable individuals and may lead to changes in prevalence rates of problem gambling \cite{gainsbury2015online}. Since individuals frequently use social media to share information about health-related issues, thereby generating large amounts of data, the use of computational methods to automatically detect risk factors or changes in prevalence is becoming increasingly important. In Germany, the legalisation of online gambling as part of the GlüStV 2021 \cite{gluestv21} may pose a special risk for vulnerable individuals. German operators of gambling halls are legally obliged to provide evidence of a social concept for gambling venues, which includes training on prevention measures and detection of addictive behaviours \cite{gluestv21}. In contrast, online gambling is harder to monitor, and users can access illegal content more easily. As individuals with problematic gambling behaviour appear to be strongly involved in online communities \cite{sirola2018excessive,sirola2019loneliness}, the study of the content produced may help to indirectly measure changes in gambling behaviour. Automatically detecting signs of problem gambling may therefore be beneficial for monitoring activities of website operators in the context of player protection on the internet. Changes in gambling behaviour have been reported during the lockdown periods following the COVID-19 pandemic. A number of studies report an increase in online gambling, especially among vulnerable individuals \cite{price2020online,sallie2021assessing}. A potential application of the model employed here could be to determine changes in problem gambling during the COVID-19 lockdown phases in Germany. Automatic detection based on online communication could be particularly beneficial when on-site measurements of gambling behaviour are not possible.

\subsection*{Conclusion}

We used ML to train a large language model to detect signs of problem gambling in online texts, finding higher prediction accuracy using manual annotation compared to previous work \cite{bucur2021early,bucur2022end,loyola2021unsl, maupome2021early}. While manual annotation is time-consuming, especially when the target items are rare, it seems to be an important requirement to produce clear, high-quality training data. In most cases, problem-gambling content was clearly distinguishable from other gambling content. Against the background of similar work \cite{bucur2021early,bucur2022end}, satisfactory classification could be obtained with less training data compared to state-of-the-art work \cite{bucur2021early,bucur2022end,loyola2021unsl,maupome2021early}. Error analysis revealed that, due to an overlap in terminology (problem- and finance-related terms), distinguishing casino complaints from problem-gambling posts posed a challenge for the model. Still, the present work shows that automatically detecting signs of problem ambling from online texts using machine learning algorithms is feasible. The results confirm the viability of using BERT when the amount of available training data is limited \cite{devlin2018bert} and demonstrate that satisfactory classification performance can be achieved by generating high-quality training material using manual annotation based on diagnostic criteria. Centrally, individual posts contain enough information to recognise signs of problem gambling, yielding potential applications of the model for preventive measures.

\section*{Data availability statement}
The raw data supporting the conclusions of this article have been scraped from a German gambling website. The data were openly accessible and technical measures designed to prevent web scraping were not disregarded. We do not hold the right to share the data, but upon request, the information and code to reproduce the analyses for scientific purposes will be shared.

\section*{Supporting information}
S1:	Annotation guide and form \\
S2:	Annotation examples

\newpage
\bibliographystyle{vancouver}

\end{document}